\documentclass{article}
\usepackage{xcolor}

\PassOptionsToPackage{numbers, sort&compress, comma, square}{natbib}
\usepackage[preprint]{neurips_2023}

\usepackage[utf8]{inputenc} 
\usepackage[T1]{fontenc}    
\usepackage{hyperref}       
\usepackage{url}            
\usepackage{booktabs}       
\usepackage{amsfonts}       
\usepackage{nicefrac}       
\usepackage{microtype}      
\usepackage{xcolor}         
\usepackage{longtable}
\usepackage{graphicx}
\usepackage{listings}
\usepackage{multirow} %
\usepackage{makecell}
\usepackage{subcaption}
\usepackage{amsmath} 
\usepackage{stfloats}
\usepackage{color}
\usepackage[most]{tcolorbox}
\usepackage{enumitem}
\usepackage{threeparttable}
\usepackage{tablefootnote}
\usepackage{marvosym}

\usepackage{float} \floatstyle{plain} \newfloat{Code}{H}{myc}

\definecolor{codegreen}{rgb}{0,0.6,0}
\definecolor{codegray}{rgb}{0.5,0.5,0.5}
\definecolor{codepurple}{rgb}{0.58,0,0.82}
\definecolor{backcolour}{rgb}{0.95,0.95,0.92}

\lstdefinestyle{mystyle}{
    backgroundcolor=\color{backcolour},   
    commentstyle=\color{codegreen},
    keywordstyle=\color{magenta},
    numberstyle=\tiny\color{codegray},
    stringstyle=\color{codepurple},
    basicstyle=\ttfamily\footnotesize,
    breakatwhitespace=false,         
    breaklines=true,                 
    captionpos=b,                    
    keepspaces=true,                 
    showspaces=false,                
    showstringspaces=false,
    showtabs=false,                  
    tabsize=2
}

\usepackage{tcolorbox}
\definecolor{lightcyan}{rgb}{0.88, 1.0, 1.0}
\colorlet{mythmback}{lightcyan!40!white}
\newtcolorbox{boxEnv}{
colback=mythmback,coltitle=blue,colframe=mythmback,
center,
width=\linewidth,
boxrule=0.5pt,
left=5pt,right=0pt,
top=2pt,bottom=2pt,
before skip=10pt, after skip=10pt
}

\title{\textit{RedStar}: Does Scaling Long-CoT Data Unlock Better Slow-Reasoning Systems?}

\author{ \quad  Haotian Xu, Xing Wu, Weinong Wang, Zhongzhi Li, Da Zheng$^{1}$    \\
      $^1$Xiaohongshu Inc\\ 
      \texttt{\{xuhaotian,wuxing,wangweinong,zhengda\}@xiaohongshu.com}\\\texttt{lizhongzhi2022@ia.ac.cn}\\
      \And
      \quad Boyuan Chen, Yi Hu, Shijia Kang, Jiaming Ji$^{2}$     \\
      $^2$Institute for Artificial Intelligence, Peking University\\ 
      \texttt{\{cbylll,huyi2002,kangshijia,jiamg.ji\}@stu.pku.edu.cn}\\
      \And
      \quad Yingying Zhang\textsuperscript{*}$^{3}$, Zhijiang Guo$^{4}$ \\
      $^3$ECNU,$^4$HKUST\\
      \texttt{yyzhang@fem.ecnu.edu.cn,zhijiangguo@hkust-gz.edu.cn}
      \And
      \quad  Yaodong Yang, Muhan Zhang$^{2}$, Debing Zhang$^{1}$\thanks{Corresponding authors.}     \\
      $^1$Xiaohongshu Inc\\ 
      $^2$Institute for Artificial Intelligence, Peking University\\ 
      \texttt{\{yaodong.yang,muhan\}@pku.edu.cn,deyang@xiaohongshu.com} \\
           }

\begin{document}
\maketitle
\newcommand{\todo}[1]{\textcolor{brown}{{[#1]}}}

\begin{abstract}
Can scaling transform reasoning? In this work, we explore the untapped potential of scaling Long Chain-of-Thought (Long-CoT) data to 1000k samples, pioneering the development of a slow-thinking model, \textit{\textbf{RedStar}}. Through extensive experiments with various LLMs and different sizes, we uncover the ingredients for specialization and scale for Long-CoT training. Surprisingly, even smaller models show significant performance gains with limited data, revealing the sample efficiency of Long-CoT and the critical role of sample difficulty in the learning process. Our findings demonstrate that Long-CoT reasoning can be effectively triggered with just a few thousand examples, while larger models achieve unparalleled improvements. We also introduce reinforcement learning (RL)-scale training as a promising direction for advancing slow-thinking systems. \textit{\textbf{RedStar}} shines across domains: on the MATH-Hard benchmark, \textit{\textbf{RedStar-code-math}} boosts performance from 66.2\% to 81.6\%, and on the USA Math Olympiad (AIME), it solves 46.7\% of problems using only 21k mixed-code-math datasets. In multimodal tasks like GeoQA and MathVista-GEO, \textit{\textbf{RedStar-Geo}} achieves competitive results with minimal Long-CoT data, outperforming other slow-thinking systems like QvQ-Preview. Compared to QwQ, \textit{\textbf{RedStar}} strikes the perfect balance between reasoning and generalizability. Our work highlights that, with careful tuning, scaling Long-CoT can unlock extraordinary reasoning capabilities-even with limited dataset and set a new standard for slow-thinking models across diverse challenges. Our data and models are released at \textcolor{blue}{\url{https://huggingface.co/RedStar-Reasoning}}.
\end{abstract}

\begin{tcolorbox}[
    colframe=orange!50!white,   
    colback=orange!5,          
    coltitle=black,           
    fonttitle=\bfseries,      
    title=Open Question For Long-COT Learning\label{long_open_q}
]
\small

\textbf{Is Slow-Thinking Reasoning Ability Benefited from Long-COT Instruction Data Scaling?} Exploring whether scaling up Long-COT instruction data enhances the model's ability to perform slow, deliberate reasoning, particularly in complex tasks. \\

\textbf{What Form of Long-COT Data Construction Has the Best Sample Efficiency?}  Investigating more effective methods for constructing Long-COT datasets to maximize reasoning performance. \\

\textbf{How Do Base Model Scale and Model Specialization Impact Long-COT Data Scaling?} Analyzing the influence of base model size (parameter count) and the volume of pretraining data on the effectiveness of Long-COT data scaling for improving reasoning capabilities. \\

\textbf{Do Long-COT and Long-COT Scale-Up Benefit Multimodal Tasks?} Examining whether Long-COT and its scaled-up versions can enhance performance in multimodal tasks that involve reasoning across text, images, and other data modalities.  
\end{tcolorbox}
\section{Introduction}

LLM-driven Slow-thinking reasoning systems, represented by models like o1\footnote{\url{https://openai.com/o1/}} and its variants—DeepSeek-R1\footnote{\url{https://api-docs.deepseek.com/news/news1120}}, k0-math\footnote{\url{https://kimi.moonshot.cn/}}, Macro-o1~\cite{zhao2024marcoo1openreasoningmodels}, and QwQ\footnote{\url{https://qwenlm.github.io/blog/qwq-32b-preview/}}—have opened up new avenues for addressing intricate reasoning challenges. At the core of these advancements is the long chain-of-thought (Long-COT) paradigm, which prioritizes methodical, step-by-step reasoning over quick, shallow responses. This approach has proven particularly effective in tasks such as mathematical problem-solving, where structured, logical steps are essential for accurate conclusions. Long-COT's success stems from its ability to break down intricate problems into smaller, manageable sub-tasks, mirroring human cognitive processes and fostering deeper, more analytical reasoning.





Recent advancements in slow-thinking reasoning systems can be categorized into tree-search-based and distillation-based methods. Tree-search approaches use data synthesis and search strategies to explore diverse reasoning paths~\cite{qin2024o1replicationjourneystrategic,zhao2024marcoo1openreasoningmodels,jiang2024technicalreportenhancingllm}, while distillation methods focus on refining performance by fine-tuning with high-quality reasoning chains~\cite{huang2024o1replicationjourney,min2024imitateexploreselfimprovereproduction,Insight-V,AtomThink}. Both approaches leverage synthetic Long-COT data to enhance models' step-by-step reasoning capabilities. However, despite their contributions, the scalability of Long-COT datasets—both in terms of volume and quality—has not been thoroughly investigated. Additionally, key factors such as multi-source data synthesis, model capacity, effective training strategies, and multimodal integration remain underexplored. These limitations hinder the broader adoption of Long-COT for optimizing reasoning performance across diverse and complex tasks.

In this paper, we investigate the potential of Long-COT from the perspective of data scaling up in  Question-Box \ref{long_open_q}. We attempt to answer whether there exists a scaling effect in constructing a large volume of Long-COT from general data, specifically, the influence of different model bases and model parameter scales on Long-COT scaling. We demonstrate that even small datasets, when properly synthesized from a moderate difficulty level, can lead to significant improvements in reasoning capabilities. Specifically, we curate a dataset consisting of 1.3k prompts and 4k prompt-response pairs, which are used to fine-tune reasoning models in a supervised manner. The results indicate that Long-COT fine-tuning yields substantial performance improvements across a range of reasoning tasks.

We first explore the impact of varying data sizes and model configurations on Long-COT scaling tuning. Our findings show that larger datasets lead to significant gains in performance, particularly in complex reasoning tasks. Smaller models, such as the \textit{7B-Instruct}, benefit substantially from larger datasets, while larger models, like the \textit{14B-Instruct}, show notable improvements even with smaller datasets. This underscores the importance of both data scaling and model capacity in optimizing reasoning performance, suggesting that tailored data sizes can enhance training efficiency depending on the model’s capabilities.

To further refine Long-COT’s effectiveness, we incorporate reinforcement learning (RL) strategies. RL-based training allows models to refine their reasoning through feedback, improving their ability to handle iterative problem-solving and fine-tuned decision-making. Our experiments demonstrate that RL techniques lead to enhanced reasoning quality, further optimizing Long-COT’s performance.

Based on our observations of base model characteristics and data scale in the Data Scaling experiments, we constructed a 4k-scale Long-COT instruction set and developed a Slow-Thinking model on the corresponding Qwen-32B-Instruct. The results show significant performance improvements, with the 32B model exhibiting better reasoning consistency and superior handling of complex tasks compared to smaller models, confirming the scalability of Long-COT methods for intricate reasoning problems. Moreover, we extend the application of Long-COT to multimodal large language models (MLLMs), addressing tasks that combine textual and visual data. By integrating step-by-step reasoning into geometric question-answering tasks, we achieve substantial improvements, demonstrating Long-COT’s versatility in both textual and multimodal contexts.

Our findings highlight the transformative potential of Long-COT for advancing reasoning models across various domains and modalities. Whether applied to mathematical problem-solving, or multimodal challenges, Long-COT tuning proves to be an essential tool for enhancing reasoning capabilities. Our final model, \textbf{\textit{RedStar}}, emerges as a leading reasoning model, demonstrating the broad impact of Long-COT methodologies.

\begin{tcolorbox}[
    colframe=orange!50!white,   
    colback=orange!5,          
    coltitle=black,           
    fonttitle=\bfseries,      
    title=Takeaways for Scaling Long-CoT Training\label{long_open_q}
]
\textbf{Long-COT enhances reasoning with limited data}: Small datasets (e.g., 1.3k prompts) can significantly improve reasoning performance, particularly in mathematical tasks, showcasing the power of Long-COT tuning even with minimal data. \\

\textbf{Larger and Specialized Models Improve Performance}: Larger models (14B, 32B) and specialized models (Math Pretraing and Context Length Extending) exhibit better performance in Long-COT training, outperforming weaker models (7B) in maintaining correct reasoning paths and handling complex tasks. \\

\textbf{Positive transfer across tasks and language}: Long-COT training boosts performance on math tasks and positively impacts other domains and language, demonstrating its versatility. Furthermore, this methodology exhibits good generalization and robustness, achieving comparable or superior results in general foundational tasks and alignment evaluation. \\

\textbf{Reinforcement learning scaling boosts efficiency}: Offline RL algorithm (DPO)  and online RL algorithm (PPO) are both effective in enhancing model performance. \\

\textbf{Long-COT strengthens multimodal models}: By adapting Long-COT methods to multimodal large language models (MLLMs), we achieve significant performance improvements, illustrating the effectiveness of slow-thinking techniques across multimodal tasks.
\end{tcolorbox}
\section{Long-COT Data Curation}
\begin{figure}[!htbp]
    \centering
    \includegraphics[width=\textwidth]{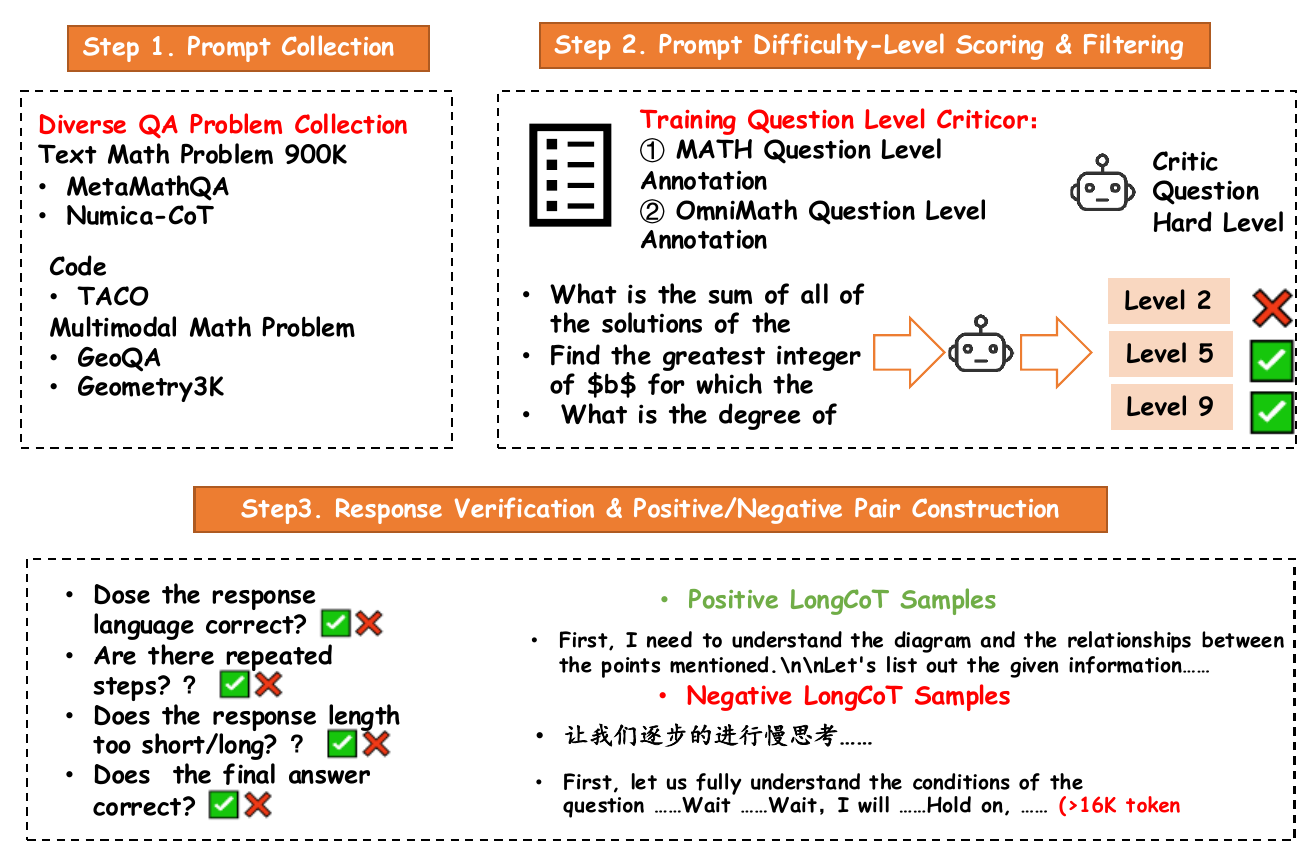}
     \caption{Overview of the Long-COT data curation process of mathematics data. The process consists of three steps: (1) collecting prompts from diverse sources, including math, code-based, and multimodal datasets; (2) annotating and filtering prompts based on difficulty levels for training; and (3) evaluating response quality and constructing positive and negative samples.}
    \label{fig:libra_guard_pipeline}
\end{figure}

In order to effectively train models on long-chain-of-thought (Long-COT) reasoning tasks, we curated a high-quality dataset that encompasses a wide range of mathematical problem-solving scenarios. The process involved data crawling, augmentation, and rigorous verification to ensure that the dataset met the specific requirements for effective reasoning at scale. This section details the methodology employed in assembling and refining the dataset used for training Long-COT models, which plays a critical role in enhancing the performance of such models in complex reasoning tasks. The following content describes the curation and augmentation of the dataset used for long-COT training, derived from the Metamath-qwen2-math dataset.

\begin{figure}[!htbp]
    \centering
    \small
    \includegraphics[width=0.8\textwidth]{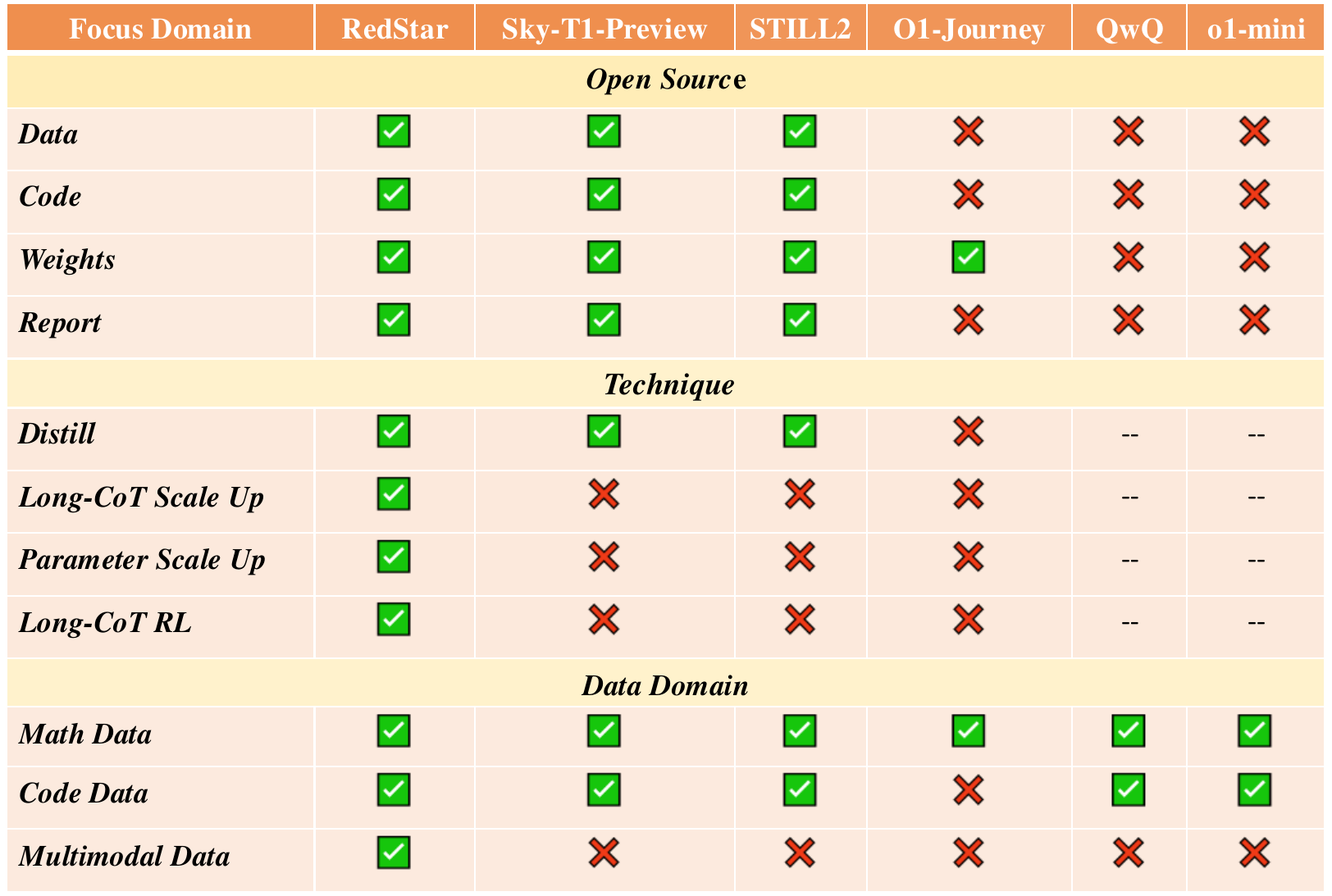}
     \caption{A comparison between our exploration and recent Slow-Thinking model approaches. We provide open-source data and code, and further investigate the impact of Long-COT on Slow-Thinking across a broader range of data domains.}
    \label{fig:libra_guard_pipeline}
\end{figure}

\paragraph{Prompt Collection} \textit{Metamath-qwen2-math Overview:} 
Metamath-qwen2-math dataset\footnote{\url{https://huggingface.co/datasets/yingyingzhang/Metamath-qwen2-math}} contains approximately 900k math problems, each with a chain-of-thought (CoT) solution. It utilizes prompts from MetaMathQA\cite{yu2024metamathbootstrapmathematicalquestions} and Numica-CoT\cite{numina_math_datasets}, with answers generated by Qwen2-Math-72-instruct\footnote{\url{https://huggingface.co/Qwen/Qwen2.5-Math-72B-Instruct}}. These answers undergo rejection sampling using the rule-evaluation-toolkit\footnote{\url{https://github.com/QwenLM/Qwen2.5-Math}} to ensure alignment with official solutions. The dataset also incorporates diverse non-synthetic data from NuminaMath-CoT, including sources like aops\_forum, amc\_aime, and cn\_k12. For the code prompt collection, we utilize problems from the TACO dataset, which comprises 25,443 training problems sourced from Codeforces, Leetcode, AtCoder, and others. Each problem comes with fine-grained labels such as algorithms, skills, and difficulty levels. The dataset also provides diverse solutions along with test cases to effectively evaluate the candidate code. For geometric scenes in multimodal tasks, we use GeoQA, which contains 3,499 question prompts. Each prompt includes geometric textual information closely related to the question and a corresponding geometric image.

\paragraph{Difficulty-Level Scoring and Augmentation} 
A difficulty-level model trained on the MATH and Omni-MATH datasets using Qwen2.5-7B-Instruct selects problems with difficulty levels 7 and above for further augmentation. These prompts are sampled multiple times using the QwQ model, with more samples generated for higher difficulty levels. Answer correctness is validated by comparing the generated responses to official answers.

\paragraph{Response Verification} 
For each question, we sent it to QwQ for 10 times sampling and obtained a large number of responses. We used manual rules to remove overly repeated, non-target languages, and Long-COTs that did not give answers in the specified format. These positive samples were retained as \textcolor{green}{\textit{Positive Samples}} for SFT, and those that failed to pass the verification were \textcolor{red}{\textit{Negative Samples}}, which can be used for subsequent RL training. For code response collection, each problem in the dataset is given to QwQ for 10 sampling attempts. Similar to math-derived data, we first clean overly repeated data and remove Long-COTs with incomplete code generation. Then, we test the code generated by QwQ using the test cases from the TACO dataset. We retain the Long-COTs that pass all test cases as the training data for SFT, ultimately resulting in 16,713 samples.

\paragraph{Final Dataset} 
The final dataset consists of 1000k samples, including 220k prompts with difficulty levels 3-9 as the whole long-cot sft-datasets. We also derive a subset including 1.3k-prompts and 4k correct prompt-response pairs with difficulty levels 7-9. Those high-quality datasets are used for supervised fine-tuning (SFT) and significantly improves the performance of models trained on Qwen2-Math-7B and Qwen2.5-32B-Instruct\footnote{Due to the computation limit, we only use 4k-subset to finetune Qwen2.5-32B-Instruct.}.

\section{Scaling Data and Models for Long-COT Efficiency} 
We investigate the impact of varying Long-COT dataset sizes and model scales on reasoning performance, focusing on mathematical tasks. The evaluation datasets include: \textbf{Math-hard}, \textbf{Olympiad-Bench}, \textbf{College\_Math}, \textbf{High\_School\_League-24}, and \textbf{AIME24}.
\begin{itemize}[left=0pt, itemsep=1pt]
    \item \textbf{Math-hard} includes the level 5 results from the MATH dataset, totaling 1.3k problems\cite{hendrycks2021measuringmathematicalproblemsolving}.
    \item \textbf{Olympiad-Bench} consists of English-language Olympiad-level math problems from Olympiad-Bench\cite{he2024olympiadbenchchallengingbenchmarkpromoting}, totaling 675 test prompts.
    \item \textbf{College\_Math} focuses on college-level math problem solving, as proposed by \cite{tang2024mathscalescalinginstructiontuning}.
    \item \textbf{High\_School\_League-24} contains 2024 Chinese high school mathematics league problems from \cite{numina_math_datasets}, used to assess out-of-distribution (OOD) performance.
    \item \textbf{AIME24} is a widely used benchmark for evaluating math problem-solving performance.
\end{itemize}


\begin{table}[h!]
\centering \small
\setlength{\tabcolsep}{4pt} 
\caption{Performance comparison of models on various benchmarks with different Long-COT configurations. \textit{Olympiad, College, and High\_school represent Olympiad-bench, College\_math, and High\_school\_league-24, respectively. AVG is the mean score across all datasets. Other tables follow the same abbreviation conventions.}}
\label{tab:model_performance}
\resizebox{\textwidth}{!}{
\begin{threeparttable}
\begin{tabular}{llcccccc}
\toprule
 \textbf{Model} & \textbf{Long-COT} & \textbf{Math-hard} & \textbf{Olympiad} & \textbf{College} & \textbf{High\_School} & \textbf{AIME24} & \textbf{AVG} \\ 
 \midrule
\multirow{6}{*}{7B-Instruct} & 0 & 52.9 & 37.0 & 38.4 & 16.7 & 10.0 & 31.0 \\ 
~& 4k & 54.4 & 41.2 & 41.6 & 10.0 & 10.0 & 31.4 \\  
~& 8k & 57.2 & 40.9 & 41.0 & 20.0 & 13.3 & 34.4 \\ 
~& 16k & 56.3 & 41.6 & 41.9 & 20.0 & 13.3 & 34.6 \\ 
~& 32k & 59.5 & 40.0 & 43.6 & 16.7 & 10.0 & 33.9 \\ 
~ & 1000k & 71.3 & 41.6 & 42.3 & 30.0 &23.3 & \textbf{41.7}\\ 
 \midrule
\multirow{6}{*}{7B-Math-Instruct\tablefootnote{Since the context length of Qwen2.5-Math-7b-Instruct is 4096, we extend the length of context to 16483 using the Position Interpolation technique with 1000k-supervised-fineunting datasets.}} & 0 & 66.6 & 45.3 & 41.9 & 23.3 & 6.7 & 36.8 \\ 
~ & 4k & 69.0 & 48.6 & 40.5 & 26.7 & 20.0 & 41.0 \\ 
~& 8k & 69.7 & 48.4 & 42.3 & 26.7 & 20.0 & 41.4 \\ 
~& 16k & 69.4 & 48.4 & 44.2 & 43.3 & 26.7 & \textbf{46.4} \\ 
~& 32k & 72.6 & 49.9 & 42.5 & 26.7 & 23.3 & 43.0 \\ 
~ & 1000k & 73.3 & 47.3 & 43.0 & 36.7 & 26.7 & 45.4 \\ 
\midrule
\multirow{6}{*}{14B-Instruct} & 0 & 59.8 & 41.3 & 41.4 & 20.0 & 16.7 & 35.8 \\ 
~ & 4k & 69.2 & 49.8 & 42.7 & 36.7 & 23.3 & 44.3 \\ 
~& 8k & 69.0 & 48.4 & 40.8 & 40.0 & 23.3 & 44.3 \\ 
~& 16k & 69.0  & 50.5 & 45.1 & 26.7 & 26.7 & 43.6 \\ 
~& 32k & 69.5 & 51.1 & 43.1 & 33.3 & 20 & 43.4 \\ 
~ & 1000k & 73.4 & 49.9 & 43.6 & 33.3 & 26.7 & \textbf{45.4} \\ 
\bottomrule
\end{tabular}
\end{threeparttable}
}
\end{table}

\subsection{Long-COT Data Scaling Performance and Efficiency Analysis}

The results in Table \ref{tab:model_performance} clearly demonstrate that increasing the scale of Long-COT data leads to significant performance improvements across various tasks. 

\textbf{\textcolor{brown}{With the incorporation of Long into the training process, the model's reasoning capabilities can achieve continuous improvement, particularly when addressing difficult and complex problems.}} For example, the performance of the \textit{7B-Math-Instruct} model improves from 36.8 to 41.0 with 4k Long-COT examples, and further increases to 45.4 when the data scale reaches 1000k examples. The \textit{14B-Instruct} model also benefits from data scaling, achieving average scores of 35.8, 44.3 across the same data configurations.

\textbf{\textcolor{brown}{An interesting pattern emerges when comparing sample efficiency across different models.}}  While the overall performance gain from 0 to 1000k examples is around 10.0 for all models, the contributions of smaller and larger datasets differ. From 0 to 4k examples, the smaller \textit{7B-Instruct} model shows minimal improvement, while the larger \textit{14B-Instruct} and \textit{7B-Math-Instruct} model achieves a significant improvement of 8.5 and 4.2.

\begin{table}[h!]
\centering \small
\setlength{\tabcolsep}{4pt} 
\caption{Performance comparison of Long-COT configurations across various prompt difficulty levels. \textit{Olympiad, College, and High\_school represent Olympiad-bench, College\_math, and High\_school\_league-24, respectively. AVG is the mean score across all datasets. Other tables follow the same abbreviation conventions. A total of 4k samples were ultimately collected for each difficulty level. We conducted experiments on \textit{Qwen2.5-14B-Instruct} and \textit{Qwen2.5-7B--Math-Instruct.}}}
\label{tab:model_performance}
\resizebox{\textwidth}{!}{
\begin{threeparttable}
\begin{tabular}{llcccccc}
\toprule
 \textbf{Model} & \textbf{Hard-Level} & \textbf{Math-hard} & \textbf{Olympiad} & \textbf{College} & \textbf{High\_School} & \textbf{AIME24} & \textbf{AVG} \\ 
\midrule
\multirow{4}{*}{7B-Math-Instruct\tablefootnote{Since the context length of Qwen2.5-Math-7b-Instruct is 4096, we first extend the length of context to 16483 using the Position Interpolation technique with 1000k-supervised-fineunting datasets.}} & \textit{Baseline} & 66.6 & 45.3 & 41.9 & 23.3 & 6.7 & 36.7 \\ 
~ & \textit{Low} & 62.7 & 43.7 & 37.3 & 20.0 & 20.0 & 36.7 \\ 
~ & \textit{Medium} & 66.2 & 46.2 & 37.3 & 20.0 & 20.0 & 37.9 \\ 
~ & \textit{High} & 69.0 & 48.6 & 40.5 & 26.7 & 20.0 & 41.0 \\ 
\midrule
\multirow{4}{*}{14B-Instruct} & \textit{Baseline} & 59.8 & 41.3 & 41.4 & 20.0 & 16.7 & 35.8 \\ 
~ & \textit{Low} & 51.1 & 35.9 & 39.0 & 16.7 & 6.7 & 29.8 \\ 
~ & \textit{Medium} & 53.7 & 38.7 & 43.3 & 30.0 & 20.0 & 37.1 \\ 
~& \textit{High} & 69.2 & 49.8 & 42.7 & 36.7 & 23.3 & 44.3 \\ 
\bottomrule
\label{tab:hard_level_aba}
\end{tabular}
\end{threeparttable}
}
\end{table}

\textbf{\textcolor{brown}{We reveal that conducting detailed long-context or math pre-training, parameter scaling can significantly enhance the training efficiency of Long-COT.}} After training on 1000K instruction data, the \textit{7B-Instruct} can achieve performance close to that of the \textit{7B-Math-Instruct} and \textit{14B-Instruct} models at the 1000K datasize. However, 7B-Math-Instruct and 14B-Instruct achieve outstanding performance directly with the use of a small amount of Meta data.

Overall, these results emphasize the importance of Long-COT data volume in enhancing a model's reasoning capabilities, especially for challenging problems. The improvements are particularly pronounced in datasets that demand complex reasoning, such as \textit{Olympiad-bench} and \textit{AIME24}, highlighting the critical role of high-quality data in boosting model performance. However, the differences in sample utilization efficiency suggest that tailoring dataset size and structure based on model capacity could further optimize training outcomes, allowing both weaker and stronger models to maximize their potential.

\subsection{Difficult Open-ended Questions Are Good Meta-data}
\textbf{\textcolor{brown}{Our findings reveal that challenging sample problems play a pivotal role in the synthesis process of Long-COT.}} Utilizing the mathematical domain as our dataset, we performed ablation experiments by categorizing the data into three distinct difficulty levels: level 3 (Low), levels 4-6 (Medium), and levels 7-9 (High). These groupings were used to synthesize prompts and conduct experiments, allowing us to observe the differential impact of problem difficulty on the effectiveness of the Long-COT synthesis.

\subsection{Model Specialization or Parameter Scaling Improves Performance}
The results further highlight that model capability plays a crucial role in enhancing performance. Larger and more specialized models consistently outperform smaller or general-purpose models, particularly in tasks that require complex reasoning. For instance, the \textit{14B-Instruct} model achieves an average performance of 45.4 with 1000k Long-COT data, outperforming the \textit{7B-Instruct} model (41.7) and closely matching the specialized \textit{7B-Math-Instruct} model (45.4) at the same data scale.

Additionally, the math-specialized \textit{7B-Math-Instruct} model consistently outperforms the general-purpose \textit{7B-Instruct}model, demonstrating the significant benefits of domain-specific pretraining. For example, on the \textit{Math-hard} dataset, the \textit{7B-Math-Instruct} model achieves 73.3, surpassing both \textit{7B-Instruct} (71.3) and the \textit{14B-Instruct} (73.4). This performance gap remains consistent even as the Long-COT dataset size increases, indicating that task-specific tuning plays a critical role in boosting a model's reasoning capabilities.

These findings emphasize that increased model capacity—whether through a larger number of parameters or specialized training—leads to better performance. This is particularly evident in datasets that demand advanced reasoning, such as \textit{High\_school\_league} and \textit{AIME24}, where larger models outperform their smaller counterparts with greater ease. Furthermore, models with domain-specific tuning exhibit more robust and consistent improvements across all benchmark categories, showcasing the importance of scale and specialization.


\section{Reinforcement Learning Brings Further Improvement}
We further explore the effectiveness of three commonly used reinforcement learning (RL) algorithms to enhance reasoning performance:

\textbf{Direct Preference Optimization (DPO):} DPO\cite{rafailov2024directpreferenceoptimizationlanguage} aligns language models with human preferences by optimizing the model directly based on preference data, bypassing the need for explicit reward modeling.

\textbf{Proximal Policy Optimization (PPO):} PPO\cite{schulman2017proximalpolicyoptimizationalgorithms} is a widely used RL algorithm that balances exploration and exploitation by adjusting policies within constrained updates.

\textbf{REINFORCE++:} REINFORCE++\cite{hu2024reinforceplusplus} improves the classic REINFORCE algorithm by reducing variance and improving convergence speed.

\textbf{Rewards of RL:} To perform RL, we propose to use rule-based RM which is used to filter out positive/negative responses in data curation. The dataset utilized in Reinforcement Learning (RL) is a subset of the Supervised Fine-Tuning (SFT) dataset.
\begin{equation}
r(x, \hat{y}, y^*) = 
\begin{cases} 
1 & \text{if } \mathsf{verifier}(\hat{y}, y^*) = \text{True} \\
0.0 & \text{if } \mathsf{verifier}(\hat{y}, y^*) = \text{False} \\
-1 & \text{if } \hat{y}  ~~ \text{doesn't contain valid answer format} 
\end{cases}
\end{equation}

The performance results for each algorithm are shown in Table \ref{tab:rl_results}, with experimental settings detailed in Table \ref{tab:experiment_settings}.

\begin{table}[!t]
\centering \small
\setlength{\tabcolsep}{4pt} 
\caption{Performance comparison of different RL algorithms across various tasks.}
\label{tab:rl_results}
\resizebox{\textwidth}{!}{
\begin{threeparttable}
\begin{tabular}{l|cccccc}
\toprule
\textbf{Model} & \textbf{Math-hard} & \textbf{Olympiad} & \textbf{College} & \textbf{High\_School} & \textbf{AIME24} & \textbf{AVG} \\ 
\midrule
14B-Instruct & 59.8 & 41.3 & 41.4 & 20.0 & 16.7 & 35.8 \\
\midrule
14B-Instruct (Long-COT-4k) & 69.2 & 49.8 & 42.7 & 36.7 & 23.3 & 44.3 \\
~~+ DPO & 69.3 & 50.8 & 46.1 & 40.0 & 30.0 & \textbf{47.2} \\
~~+ PPO & 69.9 & 50.2 & 45.2 & 44.2 & 26.7 & \textbf{47.2} \\
~~+ REINFORCE++ & 69.6 & 48.7 & 44.6 & 36.7 & 30.0 & 45.9 \\
\bottomrule
\end{tabular}
\end{threeparttable}
}
\end{table}

The results in Table \ref{tab:rl_results} highlight the varying effectiveness of the three RL algorithms:

\begin{itemize}[left=0pt, itemsep=1pt]
    \item \textbf{DPO:} Achieved the highest average performance (47.2), excelling in tasks such as \textit{college\_math} (46.1) and \textit{high\_school\_league} (40.0), demonstrating its ability to align models with human preferences.
    \item \textbf{PPO:} Showed competitive performance, particularly in \textit{math-hard} (69.9) and \textit{high\_school\_league} (44.2), achieving the same average score (47.2) as DPO but with weaker performance in \textit{AIME24} (26.7).
    \item \textbf{REINFORCE++:} Exhibited variability across tasks, performing well in \textit{AIME24} (30.0) but showing less consistency overall, resulting in an average score of 45.9.
\end{itemize}

Overall, DPO and PPO proved to be the most effective, both achieving an average score of 47.2, while REINFORCE++ demonstrated potential but would benefit from further optimization to improve task-specific consistency.


\section{Experiments}

In this section, we validate scaling methods and training strategies previously effective on smaller models, using the Qwen-32B-Instruct model at a larger scale.

\subsection{Experiment Settings}

For our experiments, we selected the Qwen-32B-Instruct model and utilized the \texttt{transformers}\footnote{\url{https://github.com/huggingface/transformers}} library for training. To mitigate memory overflow issues with the 32B model at a 16K context length, we employed context parallelism via \texttt{Deepspeed-Ulysess}\footnote{\url{https://github.com/microsoft/DeepSpeed/tree/master/blogs/deepspeed-ulysses}}, ensuring efficient training without sacrificing long-context handling.

Considering the performance gains from scaling math-derived Long-COT data from 4k to 1000k have diminished noticeably for the 14B model, for the 32B model, we limited the math-derived Long-COT data to 4k samples. To further diversify the training data, we supplemented this with a 16k code-derived Long-COT dataset. This dataset incorporates reflective prompts and filtered responses to enrich the training process.

Additionally, since DPO and PPO produced comparable results in the 14B experiment, but PPO required significantly more resources, we chose DPO as the RL training algorithm for the 32B model.

\subsection{Main Results}

\begin{table}[t]
\centering \small
\setlength{\tabcolsep}{4pt} 
\caption{Performance comparison across various reasoning tasks with different models and strategies.}
\label{tab:main_results}
\resizebox{\textwidth}{!}{
\begin{threeparttable}
\begin{tabular}{l|ccccccc}
\toprule
\textbf{Model} & \textbf{Math-hard} & \textbf{Olympiad} & \textbf{College} & \textbf{High\_School} & \textbf{AIME24} & \textbf{AVG} \\
\midrule
o1-mini & - &  65.3 & 57.8 & 63.3 &  56.7 & - \\
\midrule
o1-preview & - & - & - &  44.6 \\
\midrule
Qwen2.5-32B-Instruct & 66.2 & 44.7 & 38.8 & 23.3 & 16.7 & 37.9   \\
Qwen-QwQ-32B\tablefootnote{We run evaluation using \url{https://github.com/QwenLM/Qwen2.5-Math}.} & 81.9 & 58.4 & 41.5 & 50.0 & 50.0 & 56.4  \\
STILL-2-32B\tablefootnote{We reproduce STILL-32B using the public-5k data\cite{min2024imitateexploreselfimprovereproduction} for supervised-finetuning with the same template to ours.} & 80.6 & 59.4 & 46.7 & 40.0 & 43.3 & 54.0  \\
\midrule
RedStar-code & 69.4 & 49.5 & 32.2 & 36.7 & 30.0 & 43.6    \\
RedStar-math & 80.4 & 58.1 & 45.3 & 46.7 & 46.7& 55.4  \\
RedStar-code-math & 81.6 & 59.7 & 43.6 & 46.7 & 46.7 & 55.7   \\
\midrule
RedStar-\textit{DPO}\tablefootnote{Due to time constraints in sampling, we utilized QwQ to conduct DPO training on both positive and negative samples.} & 83.5 & 59.9 & 45.0 & 50.0 & 53.3 & 58.3 \\
\bottomrule
\end{tabular}
\end{threeparttable}
}
\end{table}

The results in Table \ref{tab:main_results} reveal several key insights:

\begin{itemize}[left=0pt, itemsep=1pt]
    \item \textbf{Comparison between Qwen and RedStar models:} The RedStar models outperform the Qwen models across most tasks, demonstrating the effectiveness of Long-COT tuning. For instance, RedStar-math achieves an average score of 54.6, which is higher than Qwen-QwQ-32B's 56.4, particularly excelling in tasks such as \textit{High\_school} (56.7) and \textit{Aime24} (36.7). On the other hand, RedStar-DPO, which combines reinforcement learning, achieves the highest average score of 58.3, indicating the benefit of incorporating advanced training techniques to improve reasoning quality, especially in more complex reasoning tasks.

    \item \textbf{Impact of Long-COT:} Long-COT tuning significantly boosts performance across multiple domains for RedStar models. RedStar-DPO, which uses both math and code Long-COT datasets, achieves the highest average score (58.3), showcasing the synergy of combining diverse datasets. Specifically, RedStar-code, trained exclusively on code Long-COT, achieves an average score of 43.6, surpassing Qwen2.5-32B-Instruct (37.9). When math datasets are integrated (RedStar-code-math), the average score increases to 55.7, demonstrating the cross-domain benefit of using Long-COT for both math and code tasks. This integration of datasets further strengthens the model’s ability to handle complex, multi-faceted reasoning scenarios.

    \item \textbf{Task-specific performance:} RedStar-DPO stands out in task-specific performance, particularly excelling in \textit{Math-hard} (83.5) and \textit{Aime24} (53.3). This reinforces the importance of specialized reinforcement learning strategies, such as DPO, for refining model reasoning in specific domains. RedStar-DPO's performance on these tasks highlights its ability to handle difficult, multi-step reasoning challenges, a task that other models, including Qwen-QwQ-32B, struggle with. This indicates that task-specific fine-tuning can further enhance a model's capabilities and make it more efficient at handling particular types of reasoning problems.

    \item \textbf{OOD performance:} In table~\ref{tab:main_ood_results}, we use the latest Chinese Graduate Entrance Mathematics Test datasets which is an OOD dataset for all models to test the OOD performance. RedStar achieves comparable results compared to closed-source apis including DeepSeek-R1, Kimi-Math and open-sourced models inculding QwQ. Since we only use the English datasets to enhence the reasoning abilities of the instruct model, it shows remarkable language transfer capabilities on Chinese OOD-tests.
\end{itemize}

These results emphasize the transformative potential of Long-COT tuning across domains. By integrating math, code, and reinforcement learning strategies, RedStar-DPO demonstrates how advanced techniques can provide substantial performance improvements in reasoning tasks. This approach lays a strong foundation for improving reasoning models across various domains.

\subsection{Impact on General Foundational Capability}

\begin{table}[!t]
\centering \small
\setlength{\tabcolsep}{4pt} 
\caption{Aggregated performance comparison across general benchmarks. Details of each category are provided in Appendix \ref{detail_general_bench}.}
\label{tab:simplified_benchmarks}
\resizebox{\textwidth}{!}{
\begin{threeparttable}
\begin{tabular}{l|ccccccc}
\toprule
\textbf{Model} & \textbf{STEM} & \textbf{Reasoning} & \textbf{CommonSense} & \textbf{Factuality} & \textbf{LongGen} & \textbf{SedarEval} & \textbf{AVG} \\
\midrule
Qwen2.5-32B-Instruct & 80.1 & 65.5 & 68.1 & 34.0 & 88.0 & 73.1 & 68.1 \\
Qwen-QwQ-32B & 71.7 & 73.0 & 60.4 & 21.0 & 81.0 & 69.7 & 62.8 \\
RedStar-code-math & 79.2 & 73.1 & 65.8 & 36.0 & 87.1 & 71.6 & 68.8 \\
\bottomrule
\end{tabular}
\end{threeparttable}
}
\end{table}

\begin{table}[!t]
\caption{OOD results: performance comparison across various o1-like models using the latest Chinese Graduate Entrance Mathematics Test datasets in \url{https://www.chinakaoyan.com/info/article/id/585019.shtml}. All results are verified by human annotaters. ``A/B'' represents the ratio of the number of problems solved under this type of problem to the total number of problems. AVG represents the ratio of the solved problems under the total problems.}
\label{tab:main_ood_results}
\centering \small
\setlength{\tabcolsep}{4pt} 
\resizebox{0.6\textwidth}{!}{
\begin{threeparttable}
\begin{tabular}{l|cccc}
\toprule
 \textbf{Model} & \textbf{Multi-Choice} & \textbf{Blank} & \textbf{Solving} &  \textbf{AVG} \\
 \midrule
Qwen-QwQ-32B & 24/29 & \textbf{16/18} & 6/13 & 76.7   \\
K1-Math & 23/29 & 14/18 & 3/13 & 66.7     \\
DeepSeek-R1 & \textbf{27/29} & 14/18 & \textbf{7/13} & \textbf{80.0}   \\
\midrule
RedStar-code-math & \textbf{27/29} &  15/18 &6/13 & \textbf{80.0}  \\
\bottomrule
\end{tabular}
\end{threeparttable}
}
\end{table}
To evaluate the impact of long-COT training on general foundational capabilities, we conducted a comprehensive comparison across multiple benchmarks covering STEM, reasoning, CommonSense, factuality, long-text generation (Hellobench~\cite{que2024hellobench}), and SedarEval~\cite{fan2024sedareval} tasks. These benchmarks provide a comprehensive view of each model’s performance in handling general tasks. Table \ref{tab:simplified_benchmarks} summarizes the aggregated scores across these categories for Qwen2.5-32B-Instruct, Qwen-QwQ-32B, and RedStar.

\begin{itemize}[left=0pt, itemsep=1pt]
    \item \textbf{Overall Performance:} RedStar achieves the highest average score (68.8), outperforming Qwen2.5-32B-Instruct (68.1) and Qwen-QwQ-32B (62.8), demonstrating its balanced capabilities across multiple domains.
    
    \item \textbf{STEM Strength:} In STEM tasks, RedStar achieves a score of 79.2, performing closely to Qwen2.5-32B-Instruct (80.1) and significantly outperforming Qwen-QwQ-32B (71.7), demonstrating its robustness in maintaining proficiency across understanding, computation, knowledge integration, and cross-disciplinary applications.
    
    \item \textbf{Reasoning Superiority:} RedStar leads in reasoning tasks with a score of 73.1, slightly surpassing Qwen-QwQ-32B (73.0) and notably outperforming Qwen2.5-32B-Instruct (65.5). This highlights RedStar’s effectiveness in complex multi-step reasoning.
    
    \item \textbf{CommonSense Performance:} RedStar demonstrates a solid performance in CommonSense tasks, scoring 65.8, surpassing Qwen-QwQ-32B (60.4) but trailing behind Qwen2.5-32B-Instruct (68.1). This suggests that RedStar still requires further improvements to maintain its common sense performance.
    
    \item \textbf{Factuality Improvements:} RedStar scores highest in factuality tasks (36.0), outperforming Qwen2.5-32B-Instruct (34.0) and Qwen-QwQ-32B (21.0), underscoring its accuracy and consistency in producing factual outputs.
    
    \item \textbf{Long Generation (Hellobench~\cite{que2024hellobench}):} RedStar achieves strong performance in long-text generation tasks, with a score of 87.1, closely matching Qwen2.5-32B-Instruct (88.0) and outperforming Qwen-QwQ-32B (81.0). This highlights its ability to generate coherent and meaningful long-form text across diverse tasks.
    
    \item \textbf{SedarEval~\cite{fan2024sedareval} Performance:} RedStar also performs competitively on the SedarEval benchmark, achieving a score of 71.6, indicating its robustness in general chat and problem-solving tasks. It surpasses Qwen-QwQ-32B (69.7) but falls slightly behind Qwen2.5-32B-Instruct (73.1).
\end{itemize}

RedStar demonstrates consistent and robust performance across various benchmarks, excelling in reasoning, factuality, and long-text generation tasks. While its STEM and SedarEval scores are competitive, further training optimizations could enhance its CommonSense performance and close the small gap in SedarEval evaluations. These results highlight RedStar’s versatility and potential as a leading model across multiple domains.


\section{One More Step: Applying Slow-Think to MLLM}

To verify whether Long-COT also provides similar benefits in multimodal reasoning scenarios, we conducted a study on geometric reasoning, widely recognized as a challenging multimodal mathematical reasoning task.

We used the GeoQA \cite{chen2021geoqa} training set with 3,499 samples, constructing multiple-choice prompts and filtering out non-reflective responses. We also excluded samples with excessively long wait times to improve model efficiency. We ultimately tested on the GeoQA test set and two out-of-distribution (OOD) geometric benchmarks: MathVista-GEO and Geometry3K \cite{lu2021inter}. In addition to the GeoQA test set, MathVista-GEO \cite{lu2023mathvista} includes geometric proof problems such as UniGeo and scenarios with rich visual elements like Geometry3K. Geometry3K, in particular, is a benchmark with relatively rich geometric illustration elements, making end-to-end solving by large models more challenging.

\begin{table}[!t]
\centering \small
\caption{Performance comparison across different models on GeoQA and related OOD benchmarks. MathVista-GEO is the geometric subset of the MathVista-testmini dataset, while for the Geometry3K dataset, we adopted the same input format as used in previous works such as LANS\cite{li2024lans}, PGPSNet\cite{zhang2023multi}, and GeoX\cite{xia2024geox}.}
\small
\begin{tabular}{l|ccc|c}
\hline
\textbf{MLLM} & \textbf{GeoQA-full} & \textbf{MathVista-GEO} & \textbf{Geometry3K} & \textbf{Instruction Data Size} \\
\hline
\multicolumn{5}{c}{\textit{SOTA API Model}} \\\hline
GPT4V & 43.4 & 51.0 & 33.7 & -- \\
GPT4o & 61.4 & 52.3 & 40.1  & -- \\
\hline
\multicolumn{5}{c}{\textit{Fast-Think Open-MLLM}} \\\hline
Intern2VL-8B & 48.6 & 65.8 & 26.5 & -- \\
Intern2VL-8B-shortcot & 52.8 & 65.2 & 29.7 & 3.5K \\
Geo-Intern2VL-8B & 71.4 & 68.9 & 30.7 & 170K \\
G-LLAVA-7B & 62.8 & 53.4 & 22.4 & 170K \\
Math-LLAVA-13B & 47.8 & 56.5 & 33.1 & 834K \\
\hline
\multicolumn{5}{c}{\textit{Slow-Thinker MLLM Model}} \\\hline
QvQ-72B-Preview & 65.0 & 81.3 & 29.4 & -- \\
RedStar-Geo-8B & 64.0 & 75.8 & 33.6 & 3.7K \\
\hline
\end{tabular}
\label{tab:MLLM_results}
\end{table}

We performed distillation based on the QwQ data, with 8 samples taken for each question, and applied a strategy similar to our pure-text approach for response validation. We selected Intern2VL-8B as the base model for Long-COT instruction fine-tuning. 

To demonstrate the advantages of LongCoT over ShortCoT in multimodal complex reasoning, we compared the results of direct fine-tuning on other large-scale multimodal geometric instruction sets. \textit{Geo-Intern2VL-8B} and \textit{G-LLAVA-7B} are trained on a dataset 30 times the size of our instruction set, which includes 170K augmented through data augmentation from GeoQA and Geometry3K. \textit{Math-LLaVA}, on the other hand, uses a multimodal ShortCoT instruction set derived from the combination of multiple image domains' CoTs, totaling over 800K instructions.
The results in Table \ref{tab:MLLM_results} demonstrate the substantial advantages of applying Long-COT to MLLMs:

\begin{itemize}[left=0pt, itemsep=1pt]
    \item \textbf{Superior Performance of RedStar-Geo-8B:} The performance of a model fine-tuned using a small set of long-CoT instructions synthesized solely from GeoQA-train surpasses that of advanced models such as GPT-4o/V, as well as fine-tuned models trained on the same dataset size and base model. 
    \item \textbf{Efficiency and Scalability:} Compared to the short CoT generated by incorporating large-scale data synthesis and enhancement strategies, the multimodal long CoT achieves a performance similar to that of Geo170K in instruction fine-tuning, despite utilizing less than 1/50th of the instruction set size.
    \item \textbf{Impact on Multimodal Reasoning:} The case study demonstrates that the MLLM model, after being trained with multimodal Long-COT, can engage in more fine-grained reflection on multimodal elements and reassess the authenticity of the multimodal conditions it generates. For out-of-distribution (OOD) datasets, such as MathVista-GEO (which includes samples from different distributions of subsets like UniGeo, Geometry3K, etc.) and the Geometry3K dataset, the multimodal long-CoT demonstrates superior generalization capability.
\end{itemize}

These findings highlight Long-COT’s pivotal role in enhancing MLLMs’ capabilities, and future work will explore its integration with other multimodal datasets to validate its scalability and efficacy.


\section{Related Work}

\paragraph{Slow-Thinking Reasoning} 

In the realm of cognitive science, it has been established that the human brain operates through two distinct functional systems: \textit{System-1} and \textit{System-2}. Recent progress in Computer Vision (CV) and Natural Language Processing (NLP) has facilitated the development of \textit{System-1} models, which are adept at rapid cognition and have been applied in a variety of straightforward and efficient methodologies. Nonetheless, tasks demanding intricate reasoning and advanced cognitive capabilities are contingent upon the deliberate and measured processes of \textit{System-2}. Slow thinking is a distinctive characteristic of the \textit{System-2} cognitive system. Unlike directly generating answers through autoregressive methods or other immediate processes, models employing slow thinking can produce outputs more deliberately and cautiously during reasoning tasks.

The \textit{Slow-Thinker LLM} has garnered widespread attention with the emergence of the o1 model and test-time scaling techniques, as compared to directly performing next-token prediction. 
Currently, various technical approaches have emerged to replicate the performance of o1, with efforts being made to establish a process for reproducing o1's performance.
Corresponding work has emerged in fields such as inference, code generation \cite{o1-Coder}, translation \cite{Macro-o1, DRT-o1}, medical reasoning \cite{Huatuo-o1}, and multimodal understanding \cite{Virgo, Insight-V}, aimed at building slow-thinking models to reduce hallucinations in language models. Advancements in slow-thinking reasoning systems focus on improving reasoning through Chain-of-Thought (CoT) prompting \cite{Huatuo-o1, g1}, reinforcement learning \cite{ReFT}, and iterative improvement \cite{ReST, ReST-EM, ReST-MCTS, Quiet-STaR, V-STaR, Expert-Iteration}.
\cite{zeng2024scaling} proposed a roadmap for O1-like systems using reinforcement learning, emphasizing policy initialization, reward design, and iterative learning, inspired by works like AlphaGo~\cite{silver2017mastering}. 
~\cite{xu2023no} proposed a Residual-based Energy Model (ResidualEBM) with Monte Carlo Tree Search (MCTS) to enhance LLMs' mathematical reasoning by ranking decision steps, improving accuracy without additional fine-tuning or human feedback. 
~\citep{zhao2024marcoo1openreasoningmodels} introduced Marco-O1, extending reasoning to open-ended tasks by integrating CoT fine-tuning with MCTS. 
~\cite{min2024imitateexploreselfimprovereproduction} developed STILL-2, a framework combining imitation, exploration, and self-improvement for training slow-thinking models, drawing from curriculum learning~\cite{bengio2009curriculum}. These works highlight the role of CoT and reinforcement learning in scaling reasoning systems and optimizing step-by-step problem-solving. 

While previous works focus on data synthesis improvements and multi-path reasoning strategies, our work diverges by exploring the key aspects of applying long-CoT tuning to build slow-thinking reasoning systems. Specifically, we explore the key roles of model and data scaling, as well as reinforcement learning, in achieving significant reasoning gains. Importantly, we further extend our research to the field of multimodal complex reasoning, represented by geometry reason\cite{li2024lans, chen2021geoqa}, demonstrating how integrating visual and textual reasoning can enhance the versatility and robustness of reasoning model training.

\paragraph{Long Context Synthesis}
Previous approaches for synthesizing long texts primarily involve concatenating shorter documents, often without mechanisms to preserve long-range dependencies. Some methods rely on random sampling and concatenation \citep{roziere2023code, chen2023longlora}, while others use KNN-based retrieval to cluster similar documents for improved coherence \citep{guu2020retrieval}. Recent works, like Quest \citep{gao2024quest}, balance semantic relevance and diversity by retrieving keyword-associated documents. However, these approaches often struggle to model dependencies across distant segments, limiting their effectiveness in long-text reasoning tasks. 
There is limited research on the synthesis of complex long reasoning paths, \cite{Huatuo-o1} constructs long CoT in the medical domain using a role-play multi-agent framework and a \textit{critic-refine} process. Our work reveals key processes in data construction within the current Slow-Thinker Reason System, focusing on data source selection and RL sample construction. It also explains the scaling effects of LongCoT in building powerful slow-think reasoning models.

\section{Conclusion and Future Work}

In this work, we explored the training of slow-thinking reasoning systems, highlighting the effectiveness of long chain-of-thought (Long-COT) tuning in improving reasoning performance with limited data. Our results show that Long-COT can be effectively triggered with minimal samples, yielding significant scale-dependent gains. We also demonstrated that Long-COT tuning does not hinder general task performance and that reinforcement learning (RL)-based training can further enhance scalability. Additionally, Long-COT methods applied to vision-language models (VLMs) resulted in notable improvements across multimodal tasks.

Future work will focus on synthesizing high-quality Long-COT datasets using only instruction-tuned (Instruct) models. This approach could increase the accessibility of Long-COT training and enhance reasoning capabilities with more scalable and automated data generation. We also plan to extend Long-COT to additional complex benchmarks to evaluate its generalization potential further.

\section*{Acknowledgments}
This research of Zhang is supported by the National Natural Science Foundation of China (NSFC grant nos. 12101241). 

\clearpage
{\small
\bibliography{nips2023}
}

\clearpage
\appendix
\section{Bitter Failure: Step-by-Step Verification}

In an effort to enhance the accuracy of long chain-of-thought (long-COT) reasoning, we introduced a step-by-step verification process to assess the correctness of the reasoning steps generated by the model. This process involves a first-error-step detection mechanism, which uses model voting to flag errors in the reasoning path\cite{zheng2024processbenchidentifyingprocesserrors}. To address the inherent challenges of verifying long-COT solutions, we proposed a distant-verification method. Initially, a compact solution is extracted from the long-COT reasoning. Then, we aplpy rejection sampling to filter out solutions with incorrect answers, followed by applying the first-error-step detection method\cite{zheng2024processbenchidentifyingprocesserrors} to identify and remove erroneous solutions. We performed five sampling iterations per solution, generating two versions of the dataset: a strict version (no detected errors) and a loose version (allowing at most two detected errors). To further analysis the relationship between critic ability and the final performance, we use Qwen2.5-14B-Instruct as weak-critic and o1-mini as strong-critic. All experiments are conducted using the Qwen2.5-32B-Instruct model with 1.3k prompts.

\begin{table}
\centering \small
\setlength{\tabcolsep}{4pt} 
\caption{Performance comparison of different verifiers (closed-source \textit{o1-mini} and open-source \textit{Qwen-14B}) on various reasoning tasks with strict and loose verification rules.}
\begin{tabular}{l|cccccc}
\hline
\textbf{Verifier}  & \textbf{Math-hard} & \textbf{Olympiad} & \textbf{College} & \textbf{High\_school} & \textbf{Aime24} & \textbf{AVG} \\\hline
32B-Instruct-no-verifier & 78.9 & 55.0 & 44.0 & 43.3 & 43.3 & 52.9 \\
\hline
+ Qwen-14B-strict-verifier & 79.2 & 56.4 & 43.1 & 50.0 & 36.7 & 53.1  \\
+ Qwen-14B-loose-verifier & 79.8 & 55.9 & 44.6 & 50.0 & 33.3 & 52.7  \\
\hline
+ o1-mini-strict-verifier & 78.8 & 56.1 & 42.5 & 40.0 & 40.0 & 51.5  \\
+ o1-mini-loose-verifier & 77.5 & 55.9 & 43.6 & 50.0 & 36.7 & 52.7  \\\hline
\end{tabular}
\label{tab:long_cot_results}
\end{table}

Despite the promising nature of the step-by-step verification process, as shown in Table~\ref{tab:long_cot_results}, the results revealed that the proposed verification method did not significantly outperform the no-verifier baseline across most tasks. The use of both strict and loose verification rules with the Qwen-14B verifier did not yield notable improvements, especially on more complex problems. While o1-mini \cite{OpenAI} is considered the best critic model for error detection in reasoning steps, it still struggled to effectively identify errors in the solutions generated by Long-COT for more challenging problems. This indicates that the current verification methods have limitations, especially when scaling to more difficult reasoning tasks. The failure to achieve substantial improvements suggests that further work is required to develop more robust and scalable oversight mechanisms that can handle the complexity of Long-COT reasoning across various domains.

The lack of significant improvement with step-by-step verification could be attributed to several factors. First, the Long-COT process itself, which breaks down problems into multiple steps, may inherently reduce the model's susceptibility to errors that verification methods are designed to detect. Additionally, the challenge of verifying multi-step reasoning in a manner that generalizes across diverse tasks remains an unresolved issue. Moreover, the sensitivity of error detection models like o1-mini to more intricate reasoning patterns may need further refinement, especially when applied to complex reasoning tasks outside the scope of its initial training.

Overall, while step-by-step verification shows promise, its current implementation did not lead to the expected gains in reasoning performance, suggesting that additional advancements in verification techniques and dataset quality are necessary to fully harness the potential of Long-COT reasoning systems.

\section{RL Experimental Setups}
The experimental configurations for each algorithm are summarized in Table \ref{tab:experiment_settings}.

\begin{table}[!h]
\centering \small
\caption{Experimental settings for DPO, PPO, and REINFORCE++ algorithms.}
\begin{tabular}{lccc}
\hline
\textbf{Parameter} & \textbf{DPO} & \textbf{PPO} & \textbf{REINFORCE++} \\
\hline
Epochs & 2 & 5 episodes & 5 episodes \\
Learning Rate & $5\times10^{-7}$ & Actor: $5\times10^{-7}$ & Actor: $5\times10^{-7}$ \\
 &  & Critic: $5\times10^{-6}$ & Critic: $5\times10^{-6}$ \\
DPO-beta & 0.3 / 0.5 & N/A & N/A \\
Learning Rate Decay & Linear & Linear & Linear \\
Train Batch Size & 64 & 1024 & 1024 \\
Sampling Temperature & N/A & 1.0 & 1.0 \\
Responses per Prompt & N/A & 2 & 2 \\
Critic Warmup Steps & N/A & 2 & 2 \\
Implementation Framework & TRL & OpenRLHF & OpenRLHF \\
\hline
\end{tabular}
\label{tab:experiment_settings}
\end{table}

\section{Performance on General Benchmarks} \label{detail_general_bench}

\subsection{STEM}

The STEM benchmarks evaluate a model's capability in science, technology, engineering, and mathematics, requiring high precision and deep understanding of technical subjects. Table \ref{tab:stem_results} provides a detailed breakdown of the performance of RedStar compared to other models on STEM-related subcategories.

\begin{table}[!t]
\centering
\caption{Performance comparison on STEM-related benchmarks.}
\small
\setlength{\tabcolsep}{2pt} 
\begin{tabular}{l|ccccccc}
\hline
\textbf{Model} & \textbf{MMLU} & \textbf{MMLU-} & \textbf{CMMU} & \textbf{CMMU-} & \textbf{CEVAL-} & \textbf{MMLU Pro} & \textbf{Average} \\
 & \textbf{~\cite{hendrycks2020mmlu}} & \textbf{Stem} & \textbf{~\cite{he2024cmmu}} & \textbf{Stem} & \textbf{Stem~\cite{huang2024ceval}} & \textbf{Engineering~\cite{wang2024mmlupro}} & \\
\hline
Qwen2.5-32B-Instruct & 83.3 & 80.1 & 85.9 & 81.7 & 83.8 & 57.4 & 80.1 \\
Qwen-QwQ-32B & 83.2 & 79.8 & 87.4 & 83.1 & 83.3 & 40.5 & 71.7 \\
RedStar-code-math & 82.7 & 79.4 & 86.8 & 82.6 & 83.7 & 48.9 & 79.2 \\
\hline
\end{tabular}
\label{tab:stem_results}
\end{table}

RedStar demonstrates robust and consistent performance across STEM tasks, achieving an average score of 79.2, which is slightly behind Qwen2.5-32B-Instruct (80.1) but significantly ahead of Qwen-QwQ-32B (71.7). In MMLU~\cite{hendrycks2020mmlu} (82.7) and CMMU~\cite{he2024cmmu} (86.8), RedStar exhibits strong capabilities, performing competitively with Qwen2.5-32B-Instruct while outperforming Qwen-QwQ-32B in most subcategories. It excels in STEM-specific tasks, including MMLU-Stem~\cite{hendrycks2020mmlu} (79.4), CMMU-Stem~\cite{he2024cmmu} (82.6), and CEVAL-Stem~\cite{huang2024ceval} (83.7), where it closely matches Qwen2.5-32B-Instruct, showcasing parity in handling advanced STEM reasoning. However, in professional engineering tasks (MMLU Pro Engineering~\cite{wang2024mmlupro}), RedStar scores 48.9, surpassing Qwen-QwQ-32B (40.5) but falling short of Qwen2.5-32B-Instruct (57.4), highlighting an area for improvement. Overall, RedStar proves to be a strong contender in STEM-related tasks.

\subsection{Reasoning}

Reasoning benchmarks evaluate the logical reasoning and problem-solving abilities of models across various challenging tasks. Table \ref{tab:reasoning_results} shows the performance of RedStar compared to other models on key reasoning benchmarks. Noting that, to ensure a robust and low variance assessment of model performance, we calculate the average over five evaluation runs for GPQA-Diamond~\cite{rein2023gpqa}

\begin{table}[!t]
\centering
\caption{Performance comparison on reasoning benchmarks.}
\small
\setlength{\tabcolsep}{4pt} 
\begin{tabular}{l|cccccc}
\hline
\textbf{Model} & \textbf{BBH} & \textbf{ARC-} & \textbf{GPQA-} & \textbf{Hella-} & \textbf{Wino-} & \textbf{Average} \\
 & \textbf{~\cite{suzgun2022challengingbbh}} & \textbf{Challenge~\cite{clark2018thinkarcc}} & \textbf{Diamond~\cite{rein2023gpqa}} & \textbf{swag~\cite{zellers2019hellaswag}} & \textbf{grande~\cite{sakaguchi2021winogrande}} & \\
\hline
Qwen2.5-32B-Instruct & 59.8 & 93.2 & 50.0 & 92.0 & 32.5 & 65.5 \\
Qwen-QwQ-32B & 79.1 & 93.6 & 53.1 & 88.1 & 51.1 & 73.0 \\
RedStar-code-math & 78.3 & 93.6 & 54.9 & 91.6 & 47.4 & 73.1 \\
\hline
\end{tabular}
\label{tab:reasoning_results}
\end{table}

RedStar demonstrates strong reasoning capabilities, achieving an average score of 73.1, which is slightly higher than Qwen-QwQ-32B (73.0) and significantly outperforms Qwen2.5-32B-Instruct (65.5).

\begin{table}[]
\caption{Performance on LiveBench Reasoning.}
\centering\small
\begin{tabular}{c|l|cccc}
\toprule
\textbf{Source}  & \textbf{Model} & \textbf{web\_of\_lies\_v2} & \textbf{zebra\_puzzle} & \textbf{spatial} & \textbf{Average} \\\midrule
\multirow{3}{*}{LiveBench official website}  & o1 & 100& 88.75& 86   & 91.6    \\
  & o1-mini  & 100     & 67   & 50   & 72.3    \\
 & Qwen-QwQ-32B   & 92 & 37.13& 44   & 57.7    \\\midrule
\multirow{3}{*}{test on open-source data\tablefootnote{We run evaluation using official code.}} & Qwen2.5-32B-Instruct & 66 & 40   & 40   & 48.7    \\
    & Qwen-QwQ-32B   & 84 & 54   & 36   & 58.0    \\
   & RedStar-code-math     & 86 &  44   & 48   & 59.3   \\\bottomrule
\end{tabular}
\label{tab:livebench_reasoning}
\end{table}

\paragraph{LiveBench Reasoning} We list the results of LiveBench reasoning in Table~\ref{tab:livebench_reasoning}.

\subsection{Common Sense}
\begin{table}[!t]
\centering \small
\setlength{\tabcolsep}{4pt} 
\caption{Performance comparison on common sense benchmarks.}
\begin{tabular}{l|ccccc}
\hline
\textbf{Model} & \textbf{TriviaQA-Wiki~\cite{joshi2017triviaqa}} & \textbf{NQ~\cite{kwiatkowski2019naturalnq}} & \textbf{Race-Middle~\cite{lai2017race}} & \textbf{Race-High~\cite{lai2017race}} & \textbf{Average} \\
\hline
Qwen2.5-32B-Instruct & 64.6 & 22.9 & 93.9 & 91.0 & 68.1 \\
Qwen-QwQ-32B & 78.6 & 31.9 & 70.2 & 61.0 & 60.4  \\
RedStar-code-math & 74.9 & 27.5 & 83.2 & 77.7 & 65.8 \\
\hline
\end{tabular}
\label{tab:commonsense_results}
\end{table}

Common sense benchmarks evaluate a model's general world knowledge and its ability to reason about everyday situations. Table \ref{tab:commonsense_results} provides a comparison of RedStar with other models on tasks that require robust understanding of common knowledge.

RedStar demonstrates balanced performance on common sense tasks, achieving an average score of 65.8, which places it between Qwen2.5-32B-Instruct (68.1) and Qwen-QwQ-32B (60.4). The model performs strongly on certain tasks while leaving room for improvement in others.


\subsection{Long Generation}

\begin{figure}[!htbp]
    \centering
    \includegraphics[width=\textwidth]{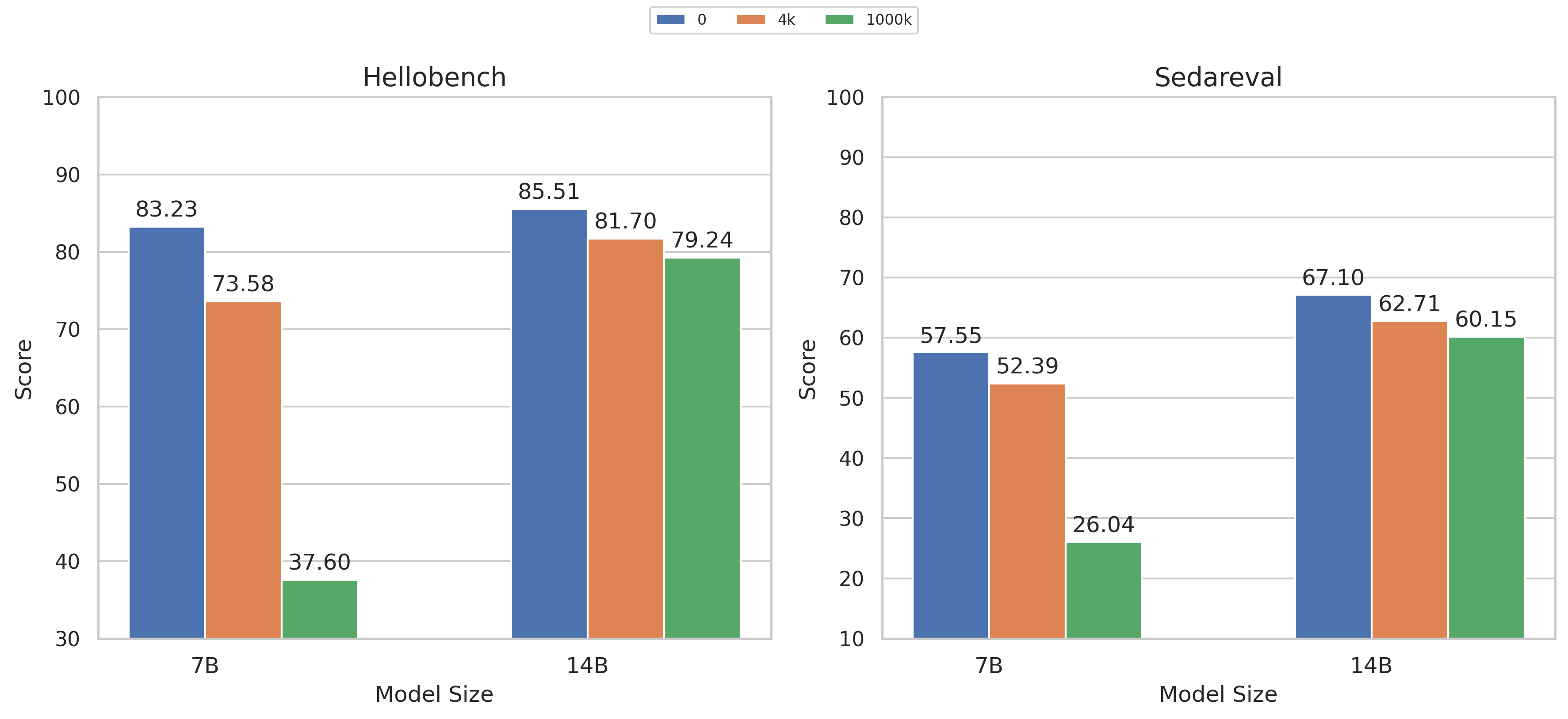}
    \caption{Results of scaling the long-cot data size and model capability on Hellobench~\cite{que2024hellobench} and Sedareval~\cite{fan2024sedareval}}
    \label{fig:general_scale}
\end{figure}

Long generation benchmarks evaluate a model's ability to generate coherent and meaningful long-form text across a variety of tasks. HelloBench~\citep{que2024hellobench} serves as a comprehensive benchmark for evaluating long-text generation capabilities, covering tasks such as open-ended QA, summarization, chat, text completion, and heuristic text generation. These tasks, based on Bloom’s Taxonomy, assess different aspects of long-context understanding and generation, offering a more holistic evaluation. The HelloEval method, introduced in HelloBench, significantly reduces the need for human evaluation while maintaining a high correlation with human judgment.

As shown in Table~\ref{tab:hellobench_results}, RedStar demonstrates strong performance across several tasks, excelling particularly in heuristic text generation, summarization, and open-ended QA. The model showcases its ability to handle complex and extended reasoning, producing coherent and relevant outputs. However, similar to other models, RedStar faces challenges when generating very long texts (over 4000 words), with issues such as repetition and quality degradation emerging. Overall, RedStar's performance on HelloBench remains competitive, particularly in shorter text generation tasks, highlighting its robust long-context reasoning capabilities.

\begin{table}[!t]
\centering \small
\setlength{\tabcolsep}{4pt} 
\caption{Performance comparison on long generation.}
\begin{tabular}{lcccccccc}
\hline
 \textbf{Model} & \textbf{Heuristic text generation} & \textbf{Summarization} & \textbf{Open-ended qa} \\\hline
Qwen2.5-32B-Instruct & 88.3 & 84.4 & 91.7 \\
Qwen-QwQ-32B & 80.2 & 81.8 & 89.3 \\
RedStar-code-math & 85.9 & 85.4 & 92.7 \\\hline
\end{tabular}

\vspace{0.5cm}

\begin{tabular}{lcccccccc}
\hline
 \textbf{Model} & \textbf{Chat} & \textbf{Text completion} & \textbf{Average wc} & \textbf{AVG} \\\hline
Qwen2.5-32B-Instruct & 85.5 & 89.8 & 1023.3 & 88.0 \\
Qwen-QwQ-32B & 75.6 & 78.1 & 2311.6 & 81.0 \\
RedStar-code-math & 82.3 & 89.2 & 2373.5 & 87.1 \\\hline
\end{tabular}
\label{tab:hellobench_results}
\end{table}

\begin{table}[!t]
\caption{Performance comparison on Sedareval.}
\centering \small
\begin{tabular}{lcccccccc}
\hline
 \textbf{Model} & \textbf{Comprehension} & \textbf{Q\&A} & \textbf{Dialogue} & \textbf{Complex Instructions} \\\hline
Qwen2.5-32B-Instruct & 68.6 & 69.8 & 69.8 & 73.1 \\
Qwen-QwQ-32B & 63.7 & 63.3 & 54.1 & 79.0 \\
RedStar-code-math & 68.0 & 69.3 & 51.3 & 78.0 \\\hline
\end{tabular}

\vspace{0.5cm}

\begin{tabular}{lcccccccc}
\hline
 \textbf{Model} & \textbf{Creation} & \textbf{Math} & \textbf{Coding} & \textbf{Reasoning} & \textbf{AVG}  \\\hline
Qwen2.5-32B-Instruct & 75.7 & 84.1 & 86.1 & 75.3 & 73.1 \\
Qwen-QwQ-32B & 72.8 & 87.4 & 93.6 & 75.6 & 69.7 \\
RedStar-code-math & 76.8 & 89.6 & 90.0 & 78.4 & 71.6  \\\hline
\end{tabular}\label{tab:sedareval_results}
\end{table}

\subsection{SedarEval Evaluation}

SedarEval~\citep{fan2024sedareval} is a novel evaluation paradigm designed to assess large language models (LLMs) using self-adaptive rubrics. Unlike traditional scoring methods, SedarEval's approach provides a more accurate reflection of the problem-solving process. The benchmark includes a wide variety of tasks, covering domains such as long-tail knowledge, mathematics, coding, and logical reasoning. Comprising 1,000 carefully curated questions, each with a detailed scoring rubric, it evaluates models on their ability to generate high-quality answers across multiple domains. The specialized evaluator language model (evaluator LM), trained on the same dataset, achieves higher concordance with human grading results compared to models like GPT-4, demonstrating its efficiency and reliability.

Our experiments show that RedStar performs exceptionally well across all tasks in SedarEval, particularly excelling in mathematical and coding tasks. Its superior factuality and reasoning abilities enable it to outperform models such as Qwen-QwQ-32B and Qwen2.5-32B-Instruct, especially on more difficult and knowledge-intensive questions. This demonstrates RedStar’s robustness and versatility in handling diverse problem-solving scenarios, making it a strong candidate for a wide range of applications.

\subsection{Instruct Following}
Table~\ref{tab:ifeval_results} shows the performance of slow-thinking models on IFEval~\cite{zhou2023instructionifeval}
dataset. Compared to Qwen2.5-32B-Instruct~\cite{qwen2.5}, slow thinking models lag significantly behind, while our model~(55.5) still surpasses the Qwen-QwQ-32B~\cite{qwq-32b-preview}(41.4). This result underscores significant potential for enhancing instruction-following ability in current slow thinking models.

\subsection{Scaling Results on General Tasks}
We investigate the impact of scaling long-cot data size on general alignment tasks, including Hellobench~\cite{que2024hellobench} and Sedareval~\cite{fan2024sedareval}. Fig.~\ref{fig:general_scale} shows that the performance of the 7B model dramatically decreases, with scores dropping from 83.23 to 37.60 on Hellobench, and from 57.55 to 26.04 on Sedareval, as long-cot data size scales from 0 to 1000k. However, the 14B model shows a steady decrease in performance. This phenomenon contrasts with the positive improvement on reason performance~(as shown in Table~\ref{tab:model_performance}). We attribute this to model capability limitations, as larger models tend to demonstrate improved stability and generalization. We hypothesize that, for small models, extensive exposure to long-cot reasoning data tends to specialize them as reasonging models, with catastrophic forgetting on other capabilities.

\begin{table}[!t]
\centering \small
\setlength{\tabcolsep}{4pt} 
\caption{Performance comparison on IFEval.}
\begin{tabular}{l|c}
\hline
\textbf{Model} & \textbf{strict prompt}  \\
\hline
Qwen2.5-32B-Instruct & 80.2  \\
Qwen-QwQ-32B & 41.4   \\
RedStar-code-math & 55.5  \\
\hline
\end{tabular}
\label{tab:ifeval_results}
\end{table}

\end{document}